\documentclass[11pt,a4paper]{article}

\usepackage[utf8]{inputenc} % allow utf-8 input
\usepackage[T1]{fontenc}    % use 8-bit T1 fonts
\usepackage{mathptmx}       % selects Times Roman as basic font
\usepackage{helvet}         % selects Helvetica as sans-serif font
\usepackage{courier}        % selects Courier as typewriter font
\usepackage{type1cm}        % activate if the above 3 fonts are

\usepackage{makeidx}         % allows index generation
\usepackage{graphicx}        % standard LaTeX graphics tool
\usepackage{multicol}        % used for the two-column index
\usepackage{multirow}
\usepackage[bottom]{footmisc}% places footnotes at page bottom

\usepackage[pdftitle={Studio Ousia's Quiz Bowl Question Answering System},pdfauthor={Ikuya Yamada, Ryuji Tamaki, Hiroyuki Shindo, Yoshiyasu Takefuji}]{hyperref}       % hyperlinks
\usepackage{url}            % simple URL typesetting
\usepackage{booktabs}       % professional-quality tables
\usepackage{amsfonts}       % blackboard math symbols
\usepackage{nicefrac}       % compact symbols for 1/2, etc.
\usepackage{tabularx}
\usepackage{amsmath}
\usepackage{microtype}      % microtypography
\usepackage{authblk}

\makeindex             % used for the subject index
\title{Studio Ousia's Quiz Bowl Question Answering System}

\author{Ikuya Yamada}
\author{Ryuji Tamaki}
\affil{Studio Ousia, 1-6-1 Otemachi Chiyoda-ku, Tokyo, Japan}
\author{Hiroyuki Shindo}
\affil{Nara Institute of Science and Technology, 8916-5 Takayama, Ikoma, Nara, Japan}
\author{Yoshiyasu Takefuji}
\affil[3]{Keio University, 5322 Endo, Fujisawa, Kanagawa, Japan}

\date{}

\begin{document}

\maketitle

\begin{abstract}
  In this chapter, we describe our question answering system, which was the winning system at the Human--Computer Question Answering (HCQA) Competition at the Thirty-first Annual Conference on Neural Information Processing Systems (NIPS).
  The competition requires participants to address a factoid question answering task referred to as \textit{quiz bowl}.
  To address this task, we use two novel neural network models and combine these models with conventional information retrieval models using a supervised machine learning model.
  Our system achieved the best performance among the systems submitted in the competition and won a match against six top human quiz experts by a wide margin.
\end{abstract}

\section{Introduction}

We present our question answering system, which was the winning solution at the Human--Computer Question Answering (HCQA) Competition held at the Thirty-first Annual Conference on Neural Information Processing Systems (NIPS) 2017.
This competition requires a system to address a unique factoid question answering (QA) task referred to as \textit{quiz bowl}, which has been studied frequently \cite{boydgraber-EtAl:2012:EMNLP-CoNLL,iyyer-EtAl:2014:EMNLP2014,iyyer-EtAl:2015,TACL1065}.
Given a question, the system is required to guess the entity that is described in the question (see Table \ref{tb:qb-example}).
One unique characteristic of this task is that the question is given one word at a time, and the system can output an answer at any time.
Moreover, the answer must be an entity that exists in Wikipedia.

To address this task, we use two neural network models and conventional information retrieval (IR) models, and we combine the outputs of these models using a supervised machine learning model.
Similar to past work \cite{iyyer-EtAl:2014:EMNLP2014,iyyer-EtAl:2015,TACL1065}, our first neural network model directly solves the task by casting it as a text classification problem.
As the entities mentioned in the question (e.g., \textit{Gregor Samsa} and \textit{The Metamorphosis} in the question shown in Table \ref{tb:qb-example}) play a significant role in guessing the answer, we use words and entities as inputs to the model.
We train the neural network model to predict the answer from a set of words and entities that appear in the question.

Given a question, our second neural network model predicts the entity types of the answer.
For example, the expected entity types of the question shown in Table \ref{tb:qb-example} are \textit{author} and \textit{person}.
We train the neural network model to predict the entity types of the answer to a question.
We adopted a convolutional neural network (CNN) \cite{kim:2014:EMNLP2014} to perform this task.

The outputs of these neural network models are used as the features of a supervised machine learning model.
We train the model with these neural-network-based features and other features including the outputs of conventional IR models.
All of these machine learning models are trained using our quiz bowl QA dataset, which was developed from two existing datasets.

Our experimental results show that the proposed approach achieved high accuracy on this task.
Furthermore, our system achieved the best performance among the systems submitted in the competition and also won a live match against six top human quiz experts by a wide margin.

\begin{table}[t]
  \caption{Example of a quiz bowl question}
  \def\arraystretch{1.4}
  \begin{tabularx}{\textwidth}{X}
  \hline\hline
  \textbf{Question:}
  The protagonist of a novel by this author is evicted from the Bridge Inn and is talked into becoming a school janitor by a character whose role is often translated as the Council Chairman. A character created by this writer is surprised to discover that he no longer likes the taste of milk, but enjoys eating rotten food. The quest for Klamm, who resides in the title structure, is taken up by K in his novel The Castle. For 10 points, name this author who wrote about Gregor Samsa being turned into an insect in ``The Metamorphosis.''\\
  \hline
  \textbf{Answer:} \textit{Franz Kafka}\\
  \hline\hline
  \end{tabularx}
  \label{tb:qb-example}
\end{table}

\section{Proposed system}

\begin{figure}[tb]
\includegraphics[width=\textwidth,clip]{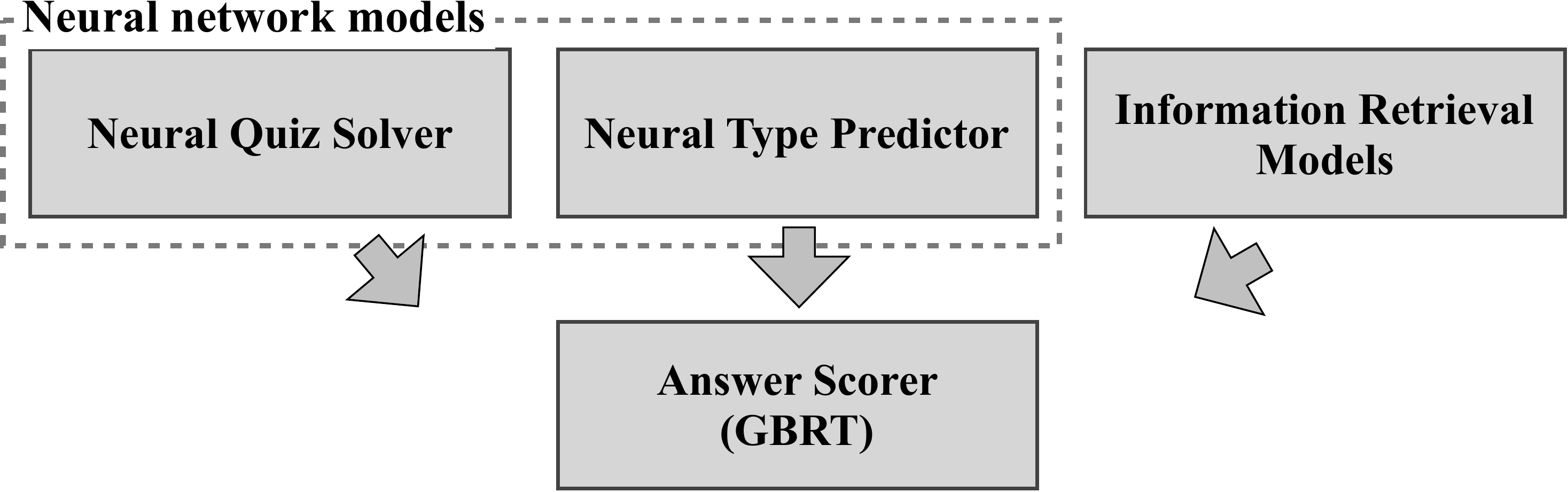}
\caption{Architecture of our proposed system.}
\label{fig:architecture}
\end{figure}

In this section, we provide an overview of the proposed system.
Figure \ref{fig:architecture} shows the architecture of our system.
We combine the outputs of two neural network models (the \textit{Neural Quiz Solver} and the \textit{Neural Type Predictor}) and conventional information retrieval (IR) models using the \textit{Answer Scorer}, which is also based on a supervised machine learning model.
We first describe the data used to develop our system and then present the technical details of our system.

\subsection{Data}
\label{subsec:data}

We used several data sources to develop our system.
First, we used the question--answer pairs available at the Protobowl website\footnote{\url{http://protobowl.com/}}, which contains over 100,000 quiz bowl questions and their answers and which was used as the official dataset of the competition.
The dataset contained several questions whose answers did not exactly match their corresponding Wikipedia titles. We resolved the answers to the corresponding Wikipedia titles using simple string matching methods and a crowd-sourcing service and excluded the questions whose answers could not be matched to Wikipedia.
In addition, we concatenated the Protobowl QA dataset with the public QA dataset provided by Iyyer et al. \cite{iyyer-EtAl:2014:EMNLP2014}, containing 20,407 quiz bowl questions and their answers.\footnote{The dataset was obtained from the authors' website: \url{https://cs.umd.edu/~miyyer/qblearn/}.}
Unlike the Protobowl dataset, the answers contained in this dataset were provided as Wikipedia titles.
Finally, we removed the duplicate questions from the concatenated dataset.
As a result, our final QA dataset contained 101,043 question--answer pairs.

We also used Wikipedia and Freebase as external data sources.
We used a Wikipedia dump generated in June 2016 and the latest Freebase data dump as obtained from the website\footnote{\url{https://developers.google.com/freebase/}}.

\subsection{Neural Quiz Solver}

We developed two neural network models to solve the QA task.
The first model is the Neural Quiz Solver, which addresses the task as a text classification problem over answers contained in the dataset.

\subsubsection{Model}

\begin{figure}[tb]
\centering
\includegraphics[width=\textwidth,clip]{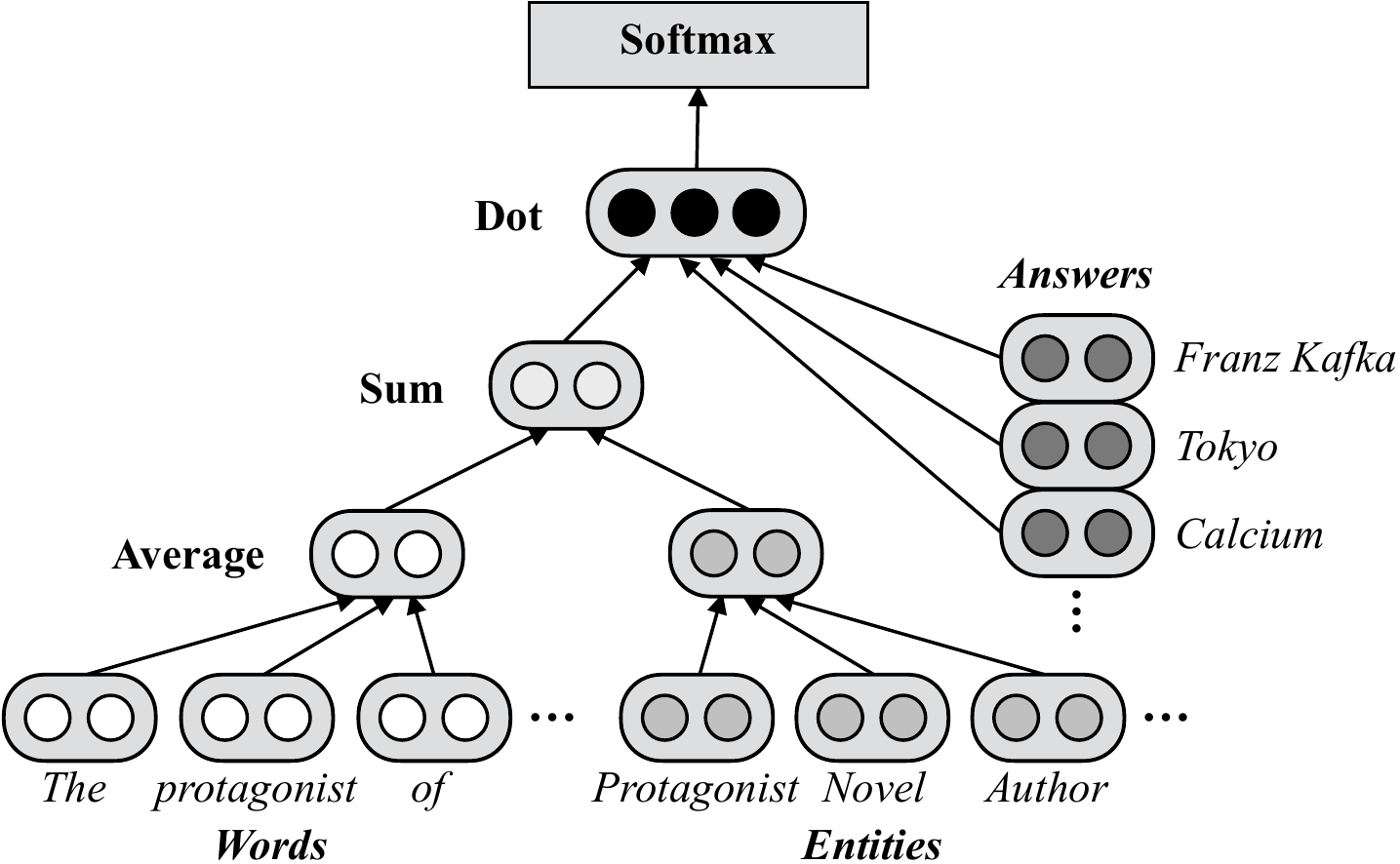}
\caption{Architecture of Neural Quiz Solver.}
\label{fig:qb_architecture}
\end{figure}

Figure \ref{fig:qb_architecture} shows the architecture of this model.
Given the words ($w_1$, $w_2$, ..., $w_N$) and the Wikipedia entities ($e_1$, $e_2$, ..., $e_K$) that appear in question $D$, our model first computes the word-based vector representation $\mathbf{v}_{D_w}$ and the entity-based vector representation $\mathbf{v}_{D_e}$ of question $D$ by averaging the vector representations of the words and the entities, respectively.
\begin{equation}
\mathbf{v}_{D_w} = \frac{1}{N}\sum_{n=1}^{N} \mathbf{W}_w\mathbf{p}_{w_n},\;
\mathbf{v}_{D_e} = \frac{1}{K}\sum_{k=1}^{K} \mathbf{W}_e\mathbf{q}_{e_k},
\label{eq:vector-averaging}
\end{equation}
where $\mathbf{p}_w \in \mathbb{R}^d$ and $\mathbf{q}_e \in \mathbb{R}^d$ are the vector representations of word $w$ and entity $e$, respectively, and $\mathbf{W}_w \in \mathbb{R}^{d\times d}$ and $\mathbf{W}_e \in \mathbb{R}^{d\times d}$ are projection matrices.
Then, the vector representation of question $\mathbf{v}_D$ is computed as the element-wise sum of $\mathbf{v}_{D_w}$ and $\mathbf{v}_{D_e}$:
\begin{equation}
\mathbf{v}_D = \mathbf{v}_{D_w} + \mathbf{v}_{D_e}
\end{equation}
Then, the probability that entity $e_t$ is the answer to the question is defined using the following softmax function:
  \begin{equation}
  \hat{y}_{e_t} = \frac{\exp(\mathbf{a}_{e_t}\!^\top \mathbf{v}_D)}{\sum_{e' \in \Gamma}\exp(\mathbf{a}_{e'}\!^\top \mathbf{v}_D)},
  \label{eq:softmax}
  \end{equation}
  where $\Gamma$ is a set containing all answers, and $\mathbf{a}_{e} \in \mathbb{R}^d$ denotes the vector representation of answer $e$.
Further, we use categorical cross entropy as a loss function.

\subsubsection{Entity detection}

Because the model requires a list of the entities appearing in a question, we automatically annotate entity names using a simple entity linking method.
The method is based on \textit{keyphraseness} \cite{Mihalcea2007}, which is the probability that an entity name is used as an anchor in Wikipedia.
We detect an entity name if its keyphraseness is larger than 2\%.
Furthermore, as an entity name can be ambiguous (e.g., \textit{Washington} can refer to the city and state in the U.S., a person's name, etc.), we use an entity name if it refers to a single entity with a probability of 95\% or more in Wikipedia.
The entities referred by the detected entity names are used as inputs to the model.

\subsubsection{Pretrained representations}

To initialize the vector representations of words ($\mathbf{p}_w$), entities ($\mathbf{q}_e$), and answers ($\mathbf{a}_e$), we use Wikipedia2Vec\footnote{\url{https://github.com/studio-ousia/wikipedia2vec}} \cite{Yamada2016}, which is our method for learning vector representations of words and entities from Wikipedia.
The model maps words and entities into the same continuous vector space; similar words and entities are placed close to one another in the vector space.

The representations of words and entities are trained by jointly optimizing the following three sub-models:
1) the conventional word-based skip-gram model, which learns to predict neighboring words given the target word in Wikipedia,
2) the \textit{anchor context model}, which learns to predict neighboring words given the target entity based on each anchor link pointing to the target entity and its context words in Wikipedia, and
3) the \textit{knowledge base graph model}, which learns to estimate neighboring entities given the target entity in the internal link graph between entities in Wikipedia.

We train the representations using the Wikipedia dump described in Section \ref{subsec:data}.
Note that we use the same pretrained entity representations to initialize the representations of entities and answers.

\subsubsection{Other details}

The model is trained by iterating over the QA dataset described in Section \ref{subsec:data}.
Because a question is given one word at a time, the model must perform accurately for incomplete questions.
To address this, we truncate a question at a random position before inputting it to the model during training.

The proposed model is implemented using PyTorch\footnote{\url{http://pytorch.org}} and trained using minibatch stochastic gradient descent (SGD) on a GPU.
The minibatch size is fixed as 32, the learning rate is automatically controlled by Adam \cite{kingma2014adam}, and the number of representation dimensions is set as $d = 300$.
We keep the parameters in the answer representations static and update all the other parameters.
To prevent overfitting, we randomly exclude the words and entities in the question with a probability of 0.5 \cite{iyyer-EtAl:2015,JMLR:v15:srivastava14a}.

Using this model, we compute two scores for each answer: (1) the predicted probability and (2) the unnormalized value inputted to the softmax function ($\mathbf{a}_{e_t}\!^\top \mathbf{v}_D$).

\subsection{Neural Type Predictor}

The second neural network model is the Neural Type Predictor, which aims to predict the entity types for a question.
For example, if the target question is the one shown in Table \ref{tb:qb-example}, the target entity types are \textit{person} and \textit{author}.
We use the FIGER entity type set \cite{Ling2012}, which consists of 112 fine-grained entity types, as the target entity types.
We automatically assign entity types to each answer by resolving the answer's Wikipedia entity to its corresponding entity in Freebase and obtaining FIGER entity types based on the mapping\footnote{The mapping was obtained from FIGER's GitHub repository: \url{https://github.com/xiaoling/figer/}.} and Freebase data.

We use two separate models with the following different target entity types: all \textit{fine-grained} entity types and only eight \textit{coarse-grained} entity types (i.e., \textit{person}, \textit{organization}, \textit{location}, \textit{product}, \textit{art}, \textit{event}, \textit{building}, and \textit{other}).
We address this task as a multiclass text classification task over entity types.
In the former setting, we address the task as a \textit{multilabel} text classification problem because many answers have multiple entity types (e.g., \textit{person} and \textit{author}).

We use a CNN \cite{kim:2014:EMNLP2014} to address this task.
Given a question consisting of a sequence of $N$ words $w_1, w_2, ..., w_N$, our task is to predict the probability for each entity type $t \in T$.
Here, a one-dimensional convolution layer of width $h \in H$ in the CNN works by moving a sliding window of size $h$ over the sequence of words.
Let the vector representation of word $w$ be $\textbf{x}_w \in \mathbb{R}^{d_{word}}$, and let the vector corresponding to the $i$-th window be
  \begin{equation}
  \mathbf{s}_i = \mathbf{x}_{w_i} \oplus \mathbf{x}_{w_{i+1}} \oplus ... \oplus \mathbf{x}_{w_{i+h-1}},
  \end{equation}
where $\oplus$ is the concatenation operator.
The result of the convolution layer consists of $m$ vectors $\mathbf{u}_1, \mathbf{u}_2, ..., \mathbf{u}_m$, each of which is computed by the following:
  \begin{equation}
  \mathbf{u}_i = \text{relu}(\mathbf{W}_{conv}\mathbf{s}_i + \mathbf{b}_{conv}),
  \end{equation}
where $\text{relu}$ is a rectifier function, $\mathbf{W}_{conv} \in \mathbb{R}^{d_{conv}\times h\cdot d_{word}}$ is a weight matrix, and $\mathbf{b}_{conv} \in \mathbb{R}^{d_{conv}}$ is a bias vector.
Note that because we use \textit{wide convolution} \cite{kalchbrenner-grefenstette-blunsom:2014:P14-1}, $m$ equals $N + h + 1$ in our model.
Then, we use max pooling to combine the $m$ vectors into a single $d_{conv}$-dimensional feature vector $\mathbf{c}$, each of whose components is computed as follows:
  \begin{equation}
  c_j = \max_{1<i\leq m} \mathbf{u}_i[j],
  \end{equation}
where $\mathbf{u}[j]$ denotes the $j$-th component of $\mathbf{u}$.
We apply multiple convolution operations with varying window sizes to obtain multiple vectors $\mathbf{c}_1, \mathbf{c}_2, ..., \mathbf{c}_{|H|}$, and obtain the concatenated feature vector $\mathbf{z} \in \mathbb{R}^{|H|\cdot d_{conv}}$ by
  \begin{equation}
  \mathbf{z} = \mathbf{c}_1 \oplus \mathbf{c}_2 \oplus ... \oplus \mathbf{c}_{|H|}.
  \end{equation}
Finally, we predict the probability corresponding to each entity type.
In the coarse-grained model, the probability corresponding to the $k$-th entity type is computed by the following softmax function:
  \begin{equation}
    \hat{y}_k = \frac{\exp(\mathbf{w}_{k}^\top \mathbf{z}  + b_{k})}{\sum_{l=1}^{|T|}\exp(\mathbf{w}_{l}^\top \mathbf{z} + b_{l})},
  \end{equation}
where $\mathbf{w}_{k} \in \mathbb{R}^{|H|\cdot d_{conv}}$ and $b_{k} \in \mathbb{R}$ are the weight vector and the bias, respectively, of the $k$-th entity type.
The model is trained to minimize categorical cross entropy.
Further, for the fine-grained model, we use the sigmoid function to create $|T|$ \textit{binary} classifiers;
the probability of the $k$-th entity type being correct is computed by
  \begin{equation}
    \hat{y}_k = \sigma(\mathbf{w}_{k}^\top \mathbf{z}  + b_{k}),
  \end{equation}
where $\sigma$ is the sigmoid function.
The model is trained to minimize binary cross entropy averaged over all entity types.

These two models are trained by iterating over our QA dataset.
We use the same configurations to train these models:
they are trained using SGD on a GPU, the minibatch size is fixed as 32, and the learning rate is controlled by Adamax \cite{kingma2014adam}.
For the hyper-parameters of the CNN, we use $H=\{2, 3, 4, 5\}$, $d_{word} = 300$, and $d_{conv} = 1,000$.
We use filter window sizes of 2, 3, 4, and 5, and 1,000 feature maps for each filter.
We use the GloVe word embeddings \cite{Pennington2014} trained on the 840 billion Common Crawl corpus to initialize the word representations.
As in the neural network model explained previously, a question is truncated at a random position before it is input to the models.
The models are implemented using PyTorch\footnote{\url{http://pytorch.org/}}.

Given a question and an answer, each model outputs two scores: the sum and the maximum probability\footnote{We aggregate probabilities because an entity can have multiple entity types in both the coarse-grained and the fine-grained models.} based on the predicted probabilities of the entity types assigned to the answer.

\subsection{Information retrieval models}

As others have in past studies \cite{iyyer-EtAl:2014:EMNLP2014,yih-EtAl:2013:ACL2013,DBLP:journals/corr/YuHBP14}, we use conventional IR models to enhance the performance of our QA system.
In particular, we compute multiple relevance scores against the documents associated with the target answer using the words in a question as a query.

Specifically, for each answer contained in the dataset, we create the target documents using the following two types of data sources: (1) \textit{Wikipedia text}, which is the page text in the answer's Wikipedia entry, and (2) \textit{dataset questions}, which are the questions contained in our QA dataset and associated with the answer.
Regarding Wikipedia text, we use two methods to create documents for each answer: treating page text as a single document and treating each paragraph as a separate document.
We also use two similar methods for dataset questions: creating a single document by concatenating all questions associated with the answer and treating each question as a separate document.
Further, because the latter methods of both data sources create multiple documents for each answer, we first compute the relevance scores for all documents and reduce them by selecting their maximum score.

We preprocess the questions and documents by converting all words to lowercase, removing stop words\footnote{We use the list of stop words contained in the scikit-learn library.}, and performing snowball stemming.
We use two scoring methods: Okapi BM25 and the number of words in common between the question and the document.
Further, we generate four types of queries for a question using (1) its words, (2) its words and bigrams, (3) its noun words, and (4) its proper noun words.\footnote{We use Apache OpenNLP to detect noun words and proper noun words.}
There are four target document sets, two scoring methods, and four query types; thus, given a question and an answer, we compute 32 relevance scores.

\subsection{Answer Scorer}
\label{subsec:learning-to-rank}

Given a question as an input, the Answer Scorer assigns a relevance score to each answer based on the outputs of the neural network models and IR models described above.
Here, we use gradient boosted regression trees (GBRT) \cite{Friedman2001}, a model that achieves state-of-the-art performance in many tasks \cite{pmlr-v14-chapelle11a,Yin:2016:RRY:2939672.2939677}.
In particular, we address the task as a binary classification problem to predict whether an answer to a given question is correct, and we use logistic loss as the loss function.

We use the probability predicted by the model as the relevance score for each answer.
Furthermore, to reduce computational cost, we assign scores only for a small number of top answer candidates.
We generate answer candidates using the union of the top five answers with the highest scores among the scores generated by the Neural Quiz Solver and the IR models.

The features used in this model are primarily based on the scores assigned by the neural network models and IR models described above.
For each score, we generate three features using (1) the score, (2) its ranking position in the answer candidates, and (3) the margin between the score and the highest score among the scores of the answer candidates.
Further, we use the following four additional features: (1) the number of words in the question, (2) the number of sentences in the question, (3) the number of FIGER entity types associated with the answer, and (4) the binary value representing whether the question contains the answer.

The model is trained using our QA dataset.
We use the GBRT implementation in LightGBM\footnote{\url{https://github.com/Microsoft/LightGBM}} with the learning rate being 0.02 and the maximum number of leaves being 400.
To maintain accuracy for incomplete questions, we generate five questions truncated at random positions per question.
One problem is that we use the same QA dataset for training both the neural network models and the target documents of the IR models; this likely causes overfitting.
To address this, we use two methods during the training of the Answer Scorer.
For the neural network models, we adopted stacked generalization \cite{wolpert1992stacked} based on 10-fold cross validation to compute scores used to train the Answer Scorer.
For the IR models, we dynamically exclude the question used to create the input query from the documents.

\section{Experiments}

In this section, we describe the experiments we conducted to evaluate the system presented in the previous section.
We first evaluated the performance of our Neural Type Predictor independently and then tested the performance of our question answering system.

\subsection{Setup}

To train and evaluate the models presented in the previous section, we used our QA dataset.
We preprocessed the dataset by excluding questions whose answers appear fewer than five times in the dataset.
Then, we randomly sampled 10\% of the questions to use as a development set and 20\% to use as a test set and used the remaining 70\% of the questions as a training set.
Thus, we obtained 49,423 training questions, 7,060 development questions, and 14,121 test questions with 5,484 unique answers.
We denote this dataset as \textit{Dataset QA}.
From this dataset, we created another dataset to train and evaluate the performance of the Neural Type Predictor by excluding questions whose answers have no entity types.
This dataset contained 39,318 training questions, 5,662 development questions, and 11,209 test questions
and is denoted as \textit{Dataset Type}.

We used the training set to train the machine learning models, the development set for early stopping (i.e., detecting the best epoch for testing), and the test set to evaluate the performance of the models.
For the IR models, we simply concatenated the training set and the development set and used this as the target documents.

We used accuracy as the performance measure of our question answering system.
To evaluate the Neural Type Predictor, we adopted different measures for the coarse-grained model and the fine-grained model.
Because the coarse-grained model addresses the task as a single-label text classification, we used accuracy as the metric of its performance,
and as the fine-grained model performs multi-label text classification, we used Precision@1, accuracy (prediction is correct if all the predicted types and no incorrect types are predicted), and F1 score (F1 score of all type predictions) as its performance metrics.
Moreover, in order to evaluate the performance for incomplete questions, we tested the models using not only the full set of sentences in a question but also the first sentence only, the first and second sentences, and the first through the third sentences.

\subsection{Results}

\begin{table}[tb]
\centering
\setlength\tabcolsep{5pt}
\caption{Results for Neural Type Predictor.}
\begin{tabular}{l|l|cccc}
\hline
Model Name                        & Metric      & Sent 1 & Sent 1--2  & Sent 1--3 & Full\\
\hline
Coarse-grained CNN                & Accuracy    & 0.95 & 0.96 & 0.97 & 0.98 \\
\hline
\multirow{3}{*}{Fine-grained CNN} & Precision@1 & 0.93 & 0.95 & 0.96 & 0.97 \\
                                  & Accuracy    & 0.56 & 0.64 & 0.69 & 0.73 \\
                                  & F1          & 0.83 & 0.87 & 0.89 & 0.91 \\
\hline
\end{tabular}
\label{tb:type-results}
\end{table}

\begin{table}[tb]
\centering
\setlength\tabcolsep{5pt}
\caption{Accuracies of our question answering system.
NQS and NTP stand for Neural Quiz Solver and Neural Type Predictor, respectively.}
\begin{tabular}{l|cccc}
\hline
Name                        & Sent 1 & Sent 1--2  & Sent 1--3 & Full \\
\hline
Full model (NQS + NTP + IR) & 0.56   & 0.78       & 0.88      & 0.97 \\
\hline
NQS                         & 0.31   & 0.54       & 0.70      & 0.88 \\
NQS + coarse-grained NTP    & 0.33   & 0.56       & 0.72      & 0.89 \\
NQS + fine-grained NTP      & 0.33   & 0.57       & 0.73      & 0.89 \\
NQS + NTP                   & 0.34   & 0.57       & 0.73      & 0.89 \\
NQS + NTP + IR-Wikipedia    & 0.48   & 0.71       & 0.84      & 0.95 \\
NQS + NTP + IR-Dataset      & 0.49   & 0.73       & 0.86      & 0.96 \\
\hline
\end{tabular}
\label{tb:qa-results}
\end{table}

Table \ref{tb:type-results} shows the performance of our Neural Type Predictor evaluated using \textit{Dataset Type}.
The coarse-grained model performed very accurately;
the accuracies exceeded 95\% for incomplete questions and 98\% for full questions.
The fine-grained model also achieved good results; its Precision@1 scores were comparable to the accuracies of the coarse-grained model.
However, the model suffered when it came to predicting all the fine-grained entity types, resulting in the relatively degraded performance in its accuracy and its F1 score.

Table \ref{tb:qa-results} shows the performance of our question answering system.
Here, we tested the performance using \textit{Dataset QA}, and used the output of the Answer Scorer to predict the answer.
Our system performed very accurately; it achieved 56\% accuracy when given only a single sentence and 97\% accuracy given the full set of sentences.
To further evaluate the effectiveness of each sub-model presented above, we added the sub-models incrementally to the Answer Scorer.
Note that the features not based on sub-models (e.g., the number of words in a question) were included in all instances.
As a result, all of the sub-models effectively contributed to the performance.
We also observed that the neural network models (i.e., Neural Quiz Solver and Neural Type Predictor) achieved good performance only for longer questions.
Further, the IR models substantially improved the performance, especially for shorter questions.

\begin{table}[tb]
\centering
\setlength\tabcolsep{7pt}
\caption{Accuracies of the top three QA systems submitted in the competition.}
\begin{tabular}{l|cc}
\hline
Name & Accuracy \\
\hline
Our system & \textbf{0.85} \\
Acelove    & 0.675 \\
Lunit.io   & 0.6 \\
Baseline   & 0.55 \\
\hline
\end{tabular}
\label{tb:accuracy}
\end{table}

\begin{figure}[tb]
\frame{\includegraphics[width=\textwidth]{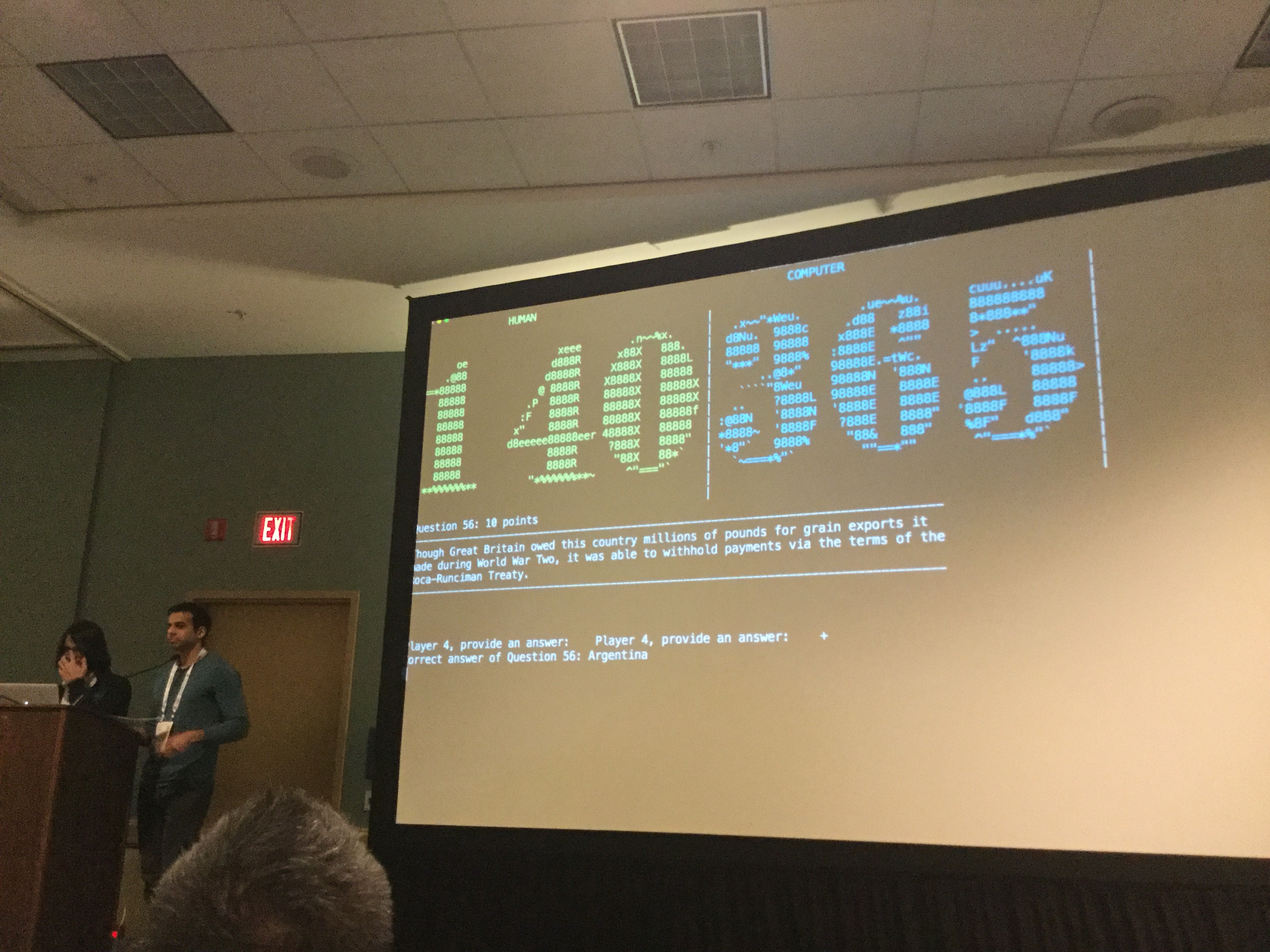}}
\caption{A live match between six top human quiz experts and our question answering system was held at the HCQA workshop at NIPS 2017.}
\label{fig:live-match}
\end{figure}

\section{Competing with other systems and human experts}

To train our final models submitted in the competition, we again used our QA dataset.
We randomly sampled 10\% of the questions as a development set and used them for early stopping.
For the IR models, we simply created the target documents using the whole dataset.

Since questions are given one word at a time, our system needed to decide whether or not to provide an answer at every word.
To achieve this, we adopted a simple strategy: we output an answer if the relevance score of the top answer exceeds a predefined threshold, which is set as 0.6.
Furthermore, as predictions frequently become unstable when the question is short, we restrict the system not to output an answer if the number of words in the question is less than 15.

Table \ref{tb:accuracy} shows the accuracies of the top three systems submitted in the competition.
Our system achieved the best performance by a wide margin.
To further evaluate the actual performance of the systems in the quiz bowl, the competition organizers performed simulated pairwise matches between the systems following the official quiz bowl rules.
Our system outperformed the Acelove system (our system: 1220 points; the Acelove system: 60 points) and the Lunit.io system (our system: 1145 points; the Lunit.io system: 105 points) by considerably wide margins.
% The experimental results are reported in detail in the past chapter summarizing the HCQA competition.

Furthermore, a live match between our system and a human team consisting of six quiz experts was held at the competition's workshop (see Figure \ref{fig:live-match}).
The human team included top quiz experts such as Raj Dhuwalia, a Jeopardy! champion and winner of 250,000 dollars on the TV show \textit{Who Wants to be a Millionaire}, and David Farris, a mathematician and three-time national champion.
Our system won the match by a significantly wide margin; it earned 425 points, whereas the human team earned only 200 points.

\section{Conclusions}

In this chapter, we describe the question answering system that we submitted in the Human--Computer Question Answering Competition held at NIPS 2017.
We proposed two novel neural network models and combined these two models with conventional IR models using a supervised machine learning model.
Our system achieved the best performance among the systems submitted in the competition and won the match against six human quiz experts by a wide margin.

\bibliography{library}
\bibliographystyle{unsrt}

\end{document}